\documentclass[conference,10pt]{IEEEtran}

\usepackage{cite}
\usepackage{graphicx}
\usepackage{amsmath}
\usepackage{algorithmic}
\usepackage[caption=false,font=footnotesize]{subfig}
\usepackage{url}
\usepackage{booktabs}

\graphicspath{
        {.}
        {diagrams/}
        {figures/}
        {images/}
}

\begin{document}
%
\title{Learning Device Models with Recurrent Neural Networks}

\author{\IEEEauthorblockN{John Clemens}
\IEEEauthorblockA{University of Maryland, Baltimore County (UMBC)\\
\texttt{clemej1@umbc.edu}}}

\maketitle

\begin{abstract}
Recurrent neural networks (RNNs) are powerful constructs capable
of modeling complex systems, up to and including Turing Machines. 
However, \textit{learning} such complex models from finite 
training sets can be difficult. In this paper we empirically show that
RNNs can learn models of computer peripheral devices through input and output 
state observation.  This enables automated development of functional 
software-only models of hardware devices.  Such models
are applicable to any number of tasks, including device validation, driver
development, code de-obfuscation, and reverse engineering.  We show that the 
same RNN structure successfully models six different devices from simple test 
circuits up to a 16550 UART serial port, and verify that these models 
are capable of producing equivalent output to real hardware.  

\end{abstract}

\section{Introduction}
In this paper we consider whether RNNs can learn functionally 
equivalent models of unknown computer hardware peripherals through 
input/output observation.  Peripheral devices attach to a main computer
and use both hardware within the device and driver software running on the
main computer to perform a task, such as printing a page or sending a message. 
However, there are instances when hardware is accessible from the main system
but driver software is not, rendering the peripheral unusable. This situation
is prevalent in open source operating systems where driver software may not 
be available from the vendor.  Without driver software or development 
documentation, it is incumbent on the system's owner to write software to make 
use of the peripheral.  The device itself is a ``black box'', with no 
information directly available to the developer beyond a set of memory 
addresses to interact with the device and the observable output of the hardware
itself. This leads to labor-intensive reverse engineering efforts with 
varying degrees of success (see e.g.~\cite{nouveau2012}).  Ideally, future 
adaptable systems should be able to automatically probe, observe, and develop
models of unknown devices to either inform the development process,
or write interface software directly.  Such solutions would also be useful
in the areas of evolutionary robotics~\cite{chen_ltl_2012} and hardware 
validation~\cite{lutsky95}~\cite{legore}.

Traditional automated black box learning techniques which learn 
\textit{exact} models, such as L*~\cite{lstar} or TTT~\cite{ttt}, are
prohibitively expensive for this task since computer peripherals have 
large input and output spaces, a large number of internal states, and require a
complex series of commands to perform a task.  Alternatively, Recurrent 
Neural Networks (RNNs) offer an intriguing solution to peripheral device 
modelling as they are able to learn \textit{approximate} models without the 
computational overhead required for traditional techniques. RNNs are versatile 
and powerful constructs that add \textit{memory} to traditional feed-forward 
neural networks via backwards (or loop) connections from output layers to 
previous layers. RNNs are capable of modeling Turing Machines~\cite{turing_rnn}.
Recent advances in machine learning hardware and software allow powerful, 
multi-layer RNNs to be trained efficiently.

The central scientific contribution of this paper is an empirical study 
of the effectiveness of using RNNs to model computer peripheral devices. 
We include a dataset of simple machines that mimic 
real device behavior (Section~\ref{sec:exp}), and then compare how well RNNs 
model those machines.  Our experiments show 
(Section~\ref{sec:results}) that RNNs are capable of learning 
functionally-equivalent models of simple hardware devices, and lay
the groundwork for future adaptable systems.


\section{Problem Statement and Assumptions}

We approach the problem of learning device models as a black box 
learning problem.  Each device $D$ has the ability to accept a sequence of
input commands $I$, such as writes to a command register or memory-mapped
addresses, which in turn produce a sequence of observable outputs $O$, such 
as data on a wire or lights on the device.  Our goal is to learn an 
approximate functional model $M$ of the device $D$ such that 
$D - M < \epsilon$, such that $\epsilon$ is minimized. 
$D - M$ is expressed in this work as the
observed functional difference (or \textit{loss}) between $D$'s response
to input $I$ and $M$'s response to the same input. 

We assume no knowledge of the inner workings of the device
being modeled. We assume that the learning algorithm either has access to a 
set of observations of the device, or has the ability
to generate such a set.  We also assume that the observed output sequence
of a device is in some way influenced by the input sequence, which we
believe is a reasonable assumption for most peripherals. Finally, we assume
that the set of observations for the device contains at least some
\textit{characteristic traces} that exercise a significant portion of the
device's capability.  These assumptions will not hold true for all potential 
scenarios, and indeed learning complete models from black box systems 
is known to be infeasible in the general case, but we assert they should
hold for a significantly large subset of target peripherals.

\section{Related Work}
\label{sec:related}

The first area of related work is the area of black box automata learning 
techniques. Black box automata learning has two main approaches: 
\textit{active} and \textit{passive} learning. Angluin~\cite{lstar} proposed
the $L^*$ active learning algorithm which can infer a Mealy machine given the 
presence of an oracle who knows the real state machine.  Variations of this 
algorithm are in use today~\cite{kearns}~\cite{ttt}, and are applied to such 
areas as hardware and software component testing~\cite{groce}, formal model 
verification~\cite{peled}, hardware reverse engineering~\cite{legore}, and 
network protocol inference~\cite{netzob}.  Recently, work has begun on learning 
\textit{register automata}, which allows a memory stack within the 
learned automata~\cite{aarts}, which may improve performance.


In passive learning, it is infeasible to actively query the device under test
so the learner is limited to the set of input and output sequences previously
gathered from the device. Gold~\cite{gold72} produced early work in this field, 
showing that inferring a minimal Moore Machine from examples was 
NP-Complete~\cite{gold78}. Passive learning algorithms descended from Gold's 
work include RPNI~\cite{rpni}, OSTIA~\cite{ostia}, and MooreMI~\cite{mooremi}.  

Automata learning approaches such as those above suffer from their need to
learn complete and accurate (though not necessarily minimal) 
representations of the systems under test. In general, the complexity of
active learning grows linearly with the number of inputs and quadratically 
with the number of states~\cite{modellearning}. Thus, active learning is only 
tractable for small problems, or large problems that can be broken down into 
smaller problems ahead of time by an expert with domain knowledge.  Passive 
learning in general is known to be NP-Complete.  The work in this paper aims
to overcome these limitations on black box algorithms by using RNNs to learn
approximate models that are close enough to the real models to be functionally
equivalent. 

Two related areas of RNN research include using RNNs or RNN-like structures
to model computing systems, such as the Neural Turing Machine~\cite{ntm} and 
Zaremba and Sutskever~\cite{zaremba2014} who train neural networks with memory 
to perform computation; and automata extraction from trained neural networks. 
Examples of extraction methods for feed-forward networks include 
FERNN~\cite{fernn}, DeepRED~\cite{deepred}, and HYPINV~\cite{hypinv}. More 
recent examples targeting RNNs include Murdoch and Szlam~\cite{relstm}, and 
Weiss, et al.~\cite{weiss_2017}. 

\section{Experimental Setup}
\label{sec:exp}

\begin{figure}
	\subfloat[EightBitMachine\label{fig:eightbitmachine}]{
		\includegraphics[width=.40\linewidth]{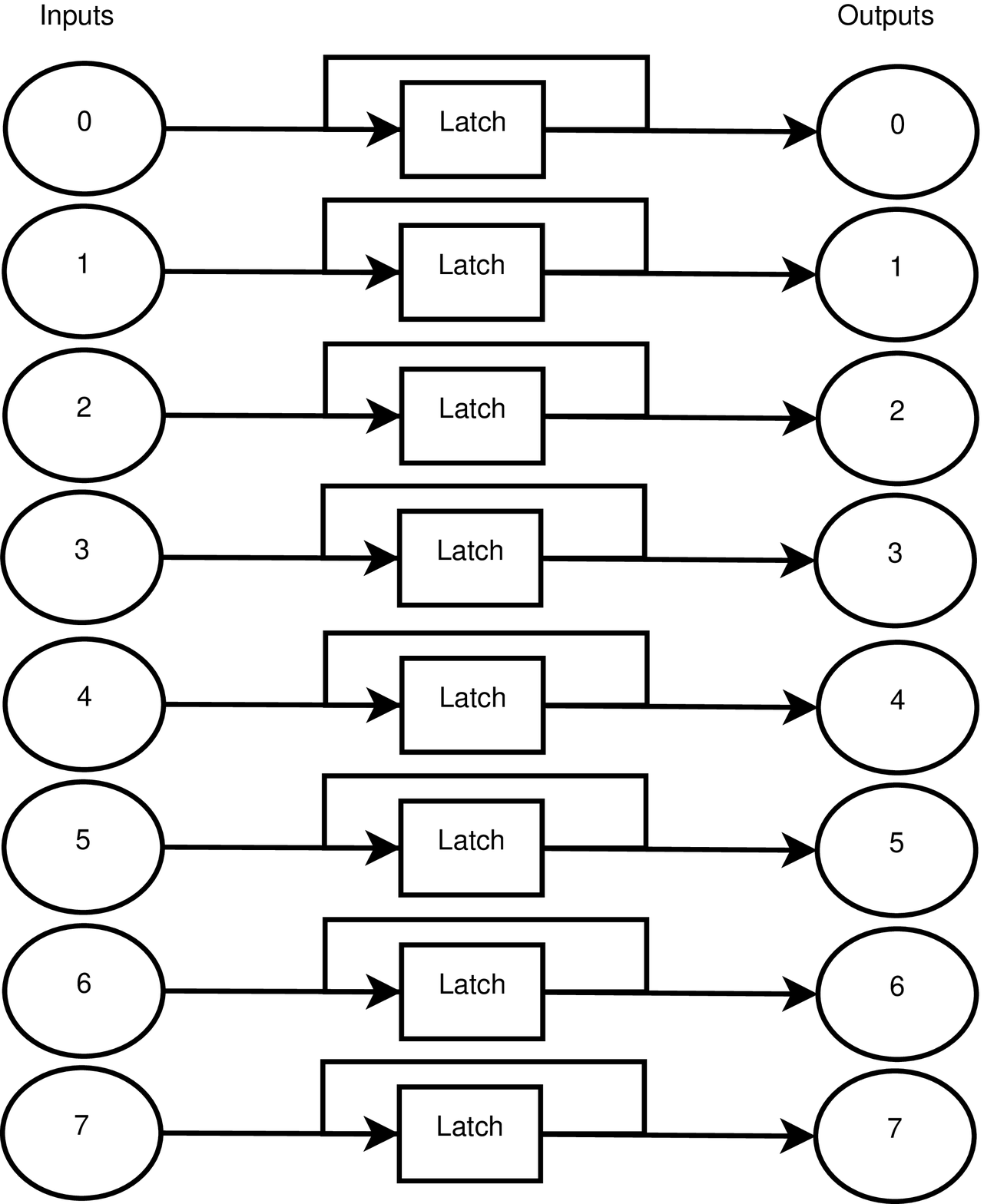}
	}
	\hfill
	\subfloat[SingleDirectMachine\label{fig:singledirectmachine}]{
		\includegraphics[width=.40\linewidth]{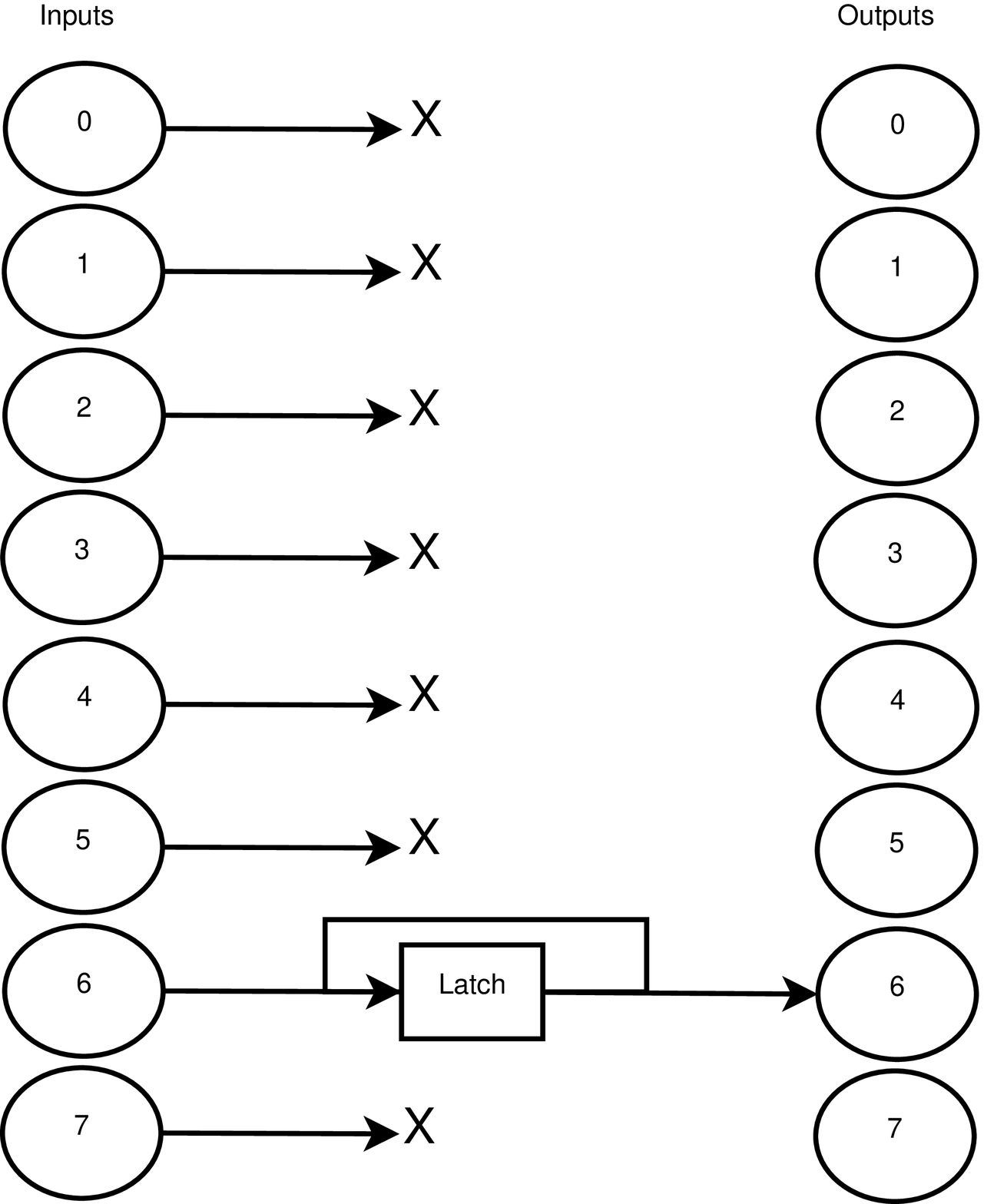}
	}
	\hfill
	\subfloat[SingleInvertMachine\label{fig:singledirectinvertmachine}]{
		\includegraphics[width=.40\linewidth]{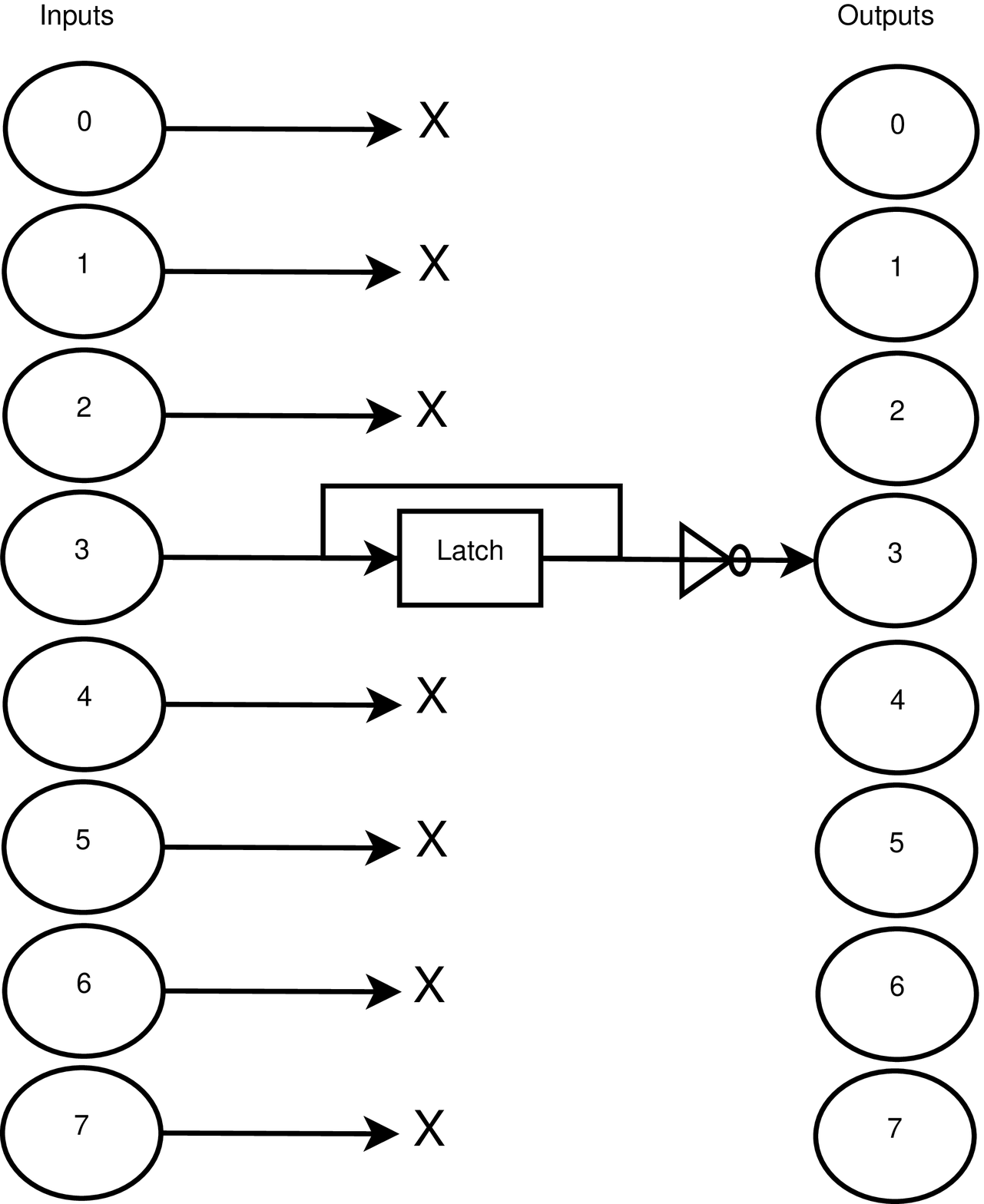}
	}
	\hfill
	\subfloat[SimpleXORMachine\label{fig:simplexormachine}]{
		\includegraphics[width=.40\linewidth]{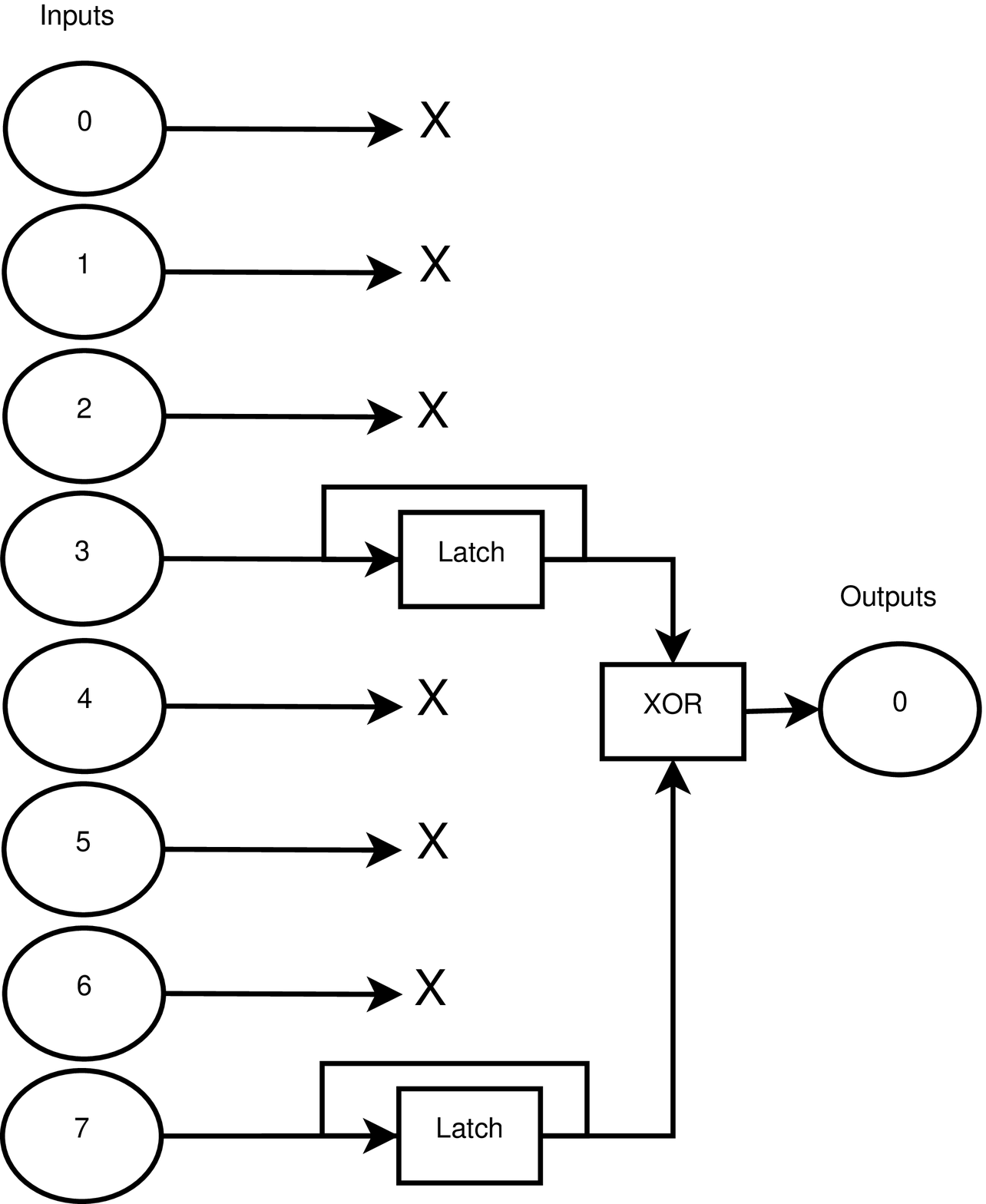}
	}
	\hfill
	\subfloat[ParityMachine\label{fig:paritymachine}]{
		\includegraphics[width=.45\linewidth]{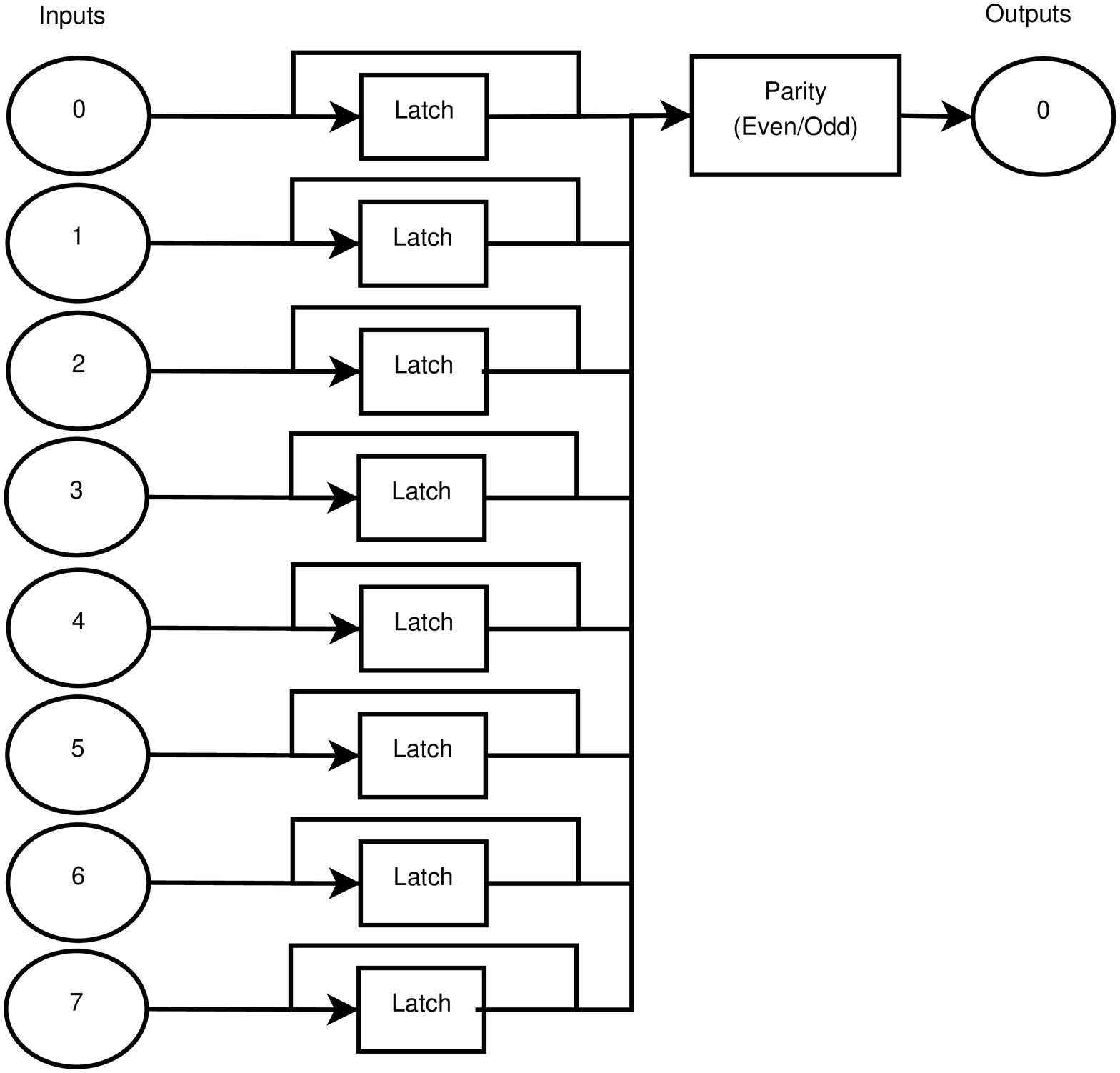}
	}
	\hfill
	\subfloat[SerialPortMachine\label{fig:serialportmachine}]{
		\includegraphics[width=.40\linewidth]{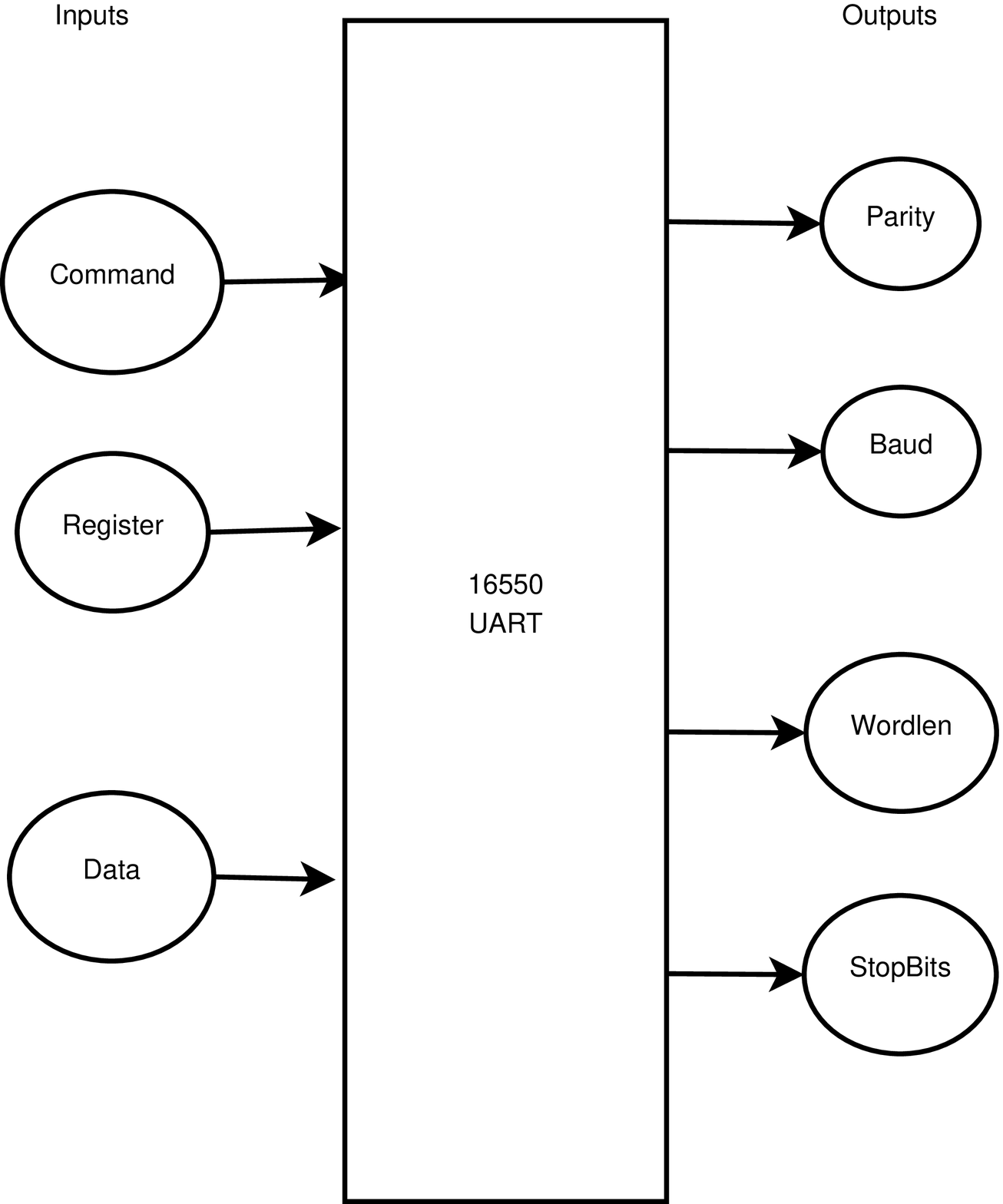}
	}

	\caption[Test Machine Models]{Simulated machines of increasing 
                complexity used for testing. All inputs are latched, meaning
                they retain the latest written value for all subsequent 
                commands.}
	\label{fig:machines}
\end{figure}

To test the ability of RNNs to learn models from devices, 
we created a set of simulated devices that perform actions which mimic 
those of real hardware peripherals\footnote{Future work will target
real hardware devices.}. The simulated devices contain interesting state 
transitions that test how well an RNN is able to learn complex concepts 
yet are simple enough for manual verification of the results.  These simulated
test devices are shown in Figure~\ref{fig:machines}. 

\subsection{Simple Machines}
\label{subsec:simplemachines}

We define a set of five simple artificial machines described below in order 
of increasing complexity.  In each machine the inputs are \textit{latched}, 
meaning that setting an input to a value of $1$ or $0$ can have an effect on 
future outputs depending on the internal structure of the machine.  In these
machines, the input value is stored in \textit{memory}, and remains the same 
for future commands in the sequence until explicitly changed.

\begin{itemize}
	\item{\textbf{EightBitMachine}: A simple mapping of 8 inputs to 8 outputs.}
	\item{\textbf{SingleDirectMachine}: 7 inputs are ignored
		and one leads directly to a single output.}
	\item{\textbf{SingleInvertMachine}: Same as a above, but output is inverted.}
	\item{\textbf{SimpleXORMachine}: 6 inputs are ignored, and the
		remaining two are XOR'd together to produce the single 
		output.}
	\item{\textbf{ParityMachine}: The output is set to $1$ if an odd number of 
                inputs are set, and $0$ if an even number are set.}
\end{itemize}

An evaluation of the magnitude of the state spaces of these machines is 
available in Appendix~\ref{app:statespace}.  These simple models provide an 
easily decomposable testing ground to test what RNNs can learn about a machine.
Success in modeling these devices will provide confidence that the RNN will be 
able to handle more complex systems in the future. 

\subsection{16550 UART}
\label{subsec:16550}

While the simple models detailed above test the basic ability of RNNs
to learn discrete concepts used by devices, the 16550 UART model shows a 
practical application of the RNN approach to a real world peripheral. The 
16550 UART~\cite{ns16550} is the component behind the modern PC serial port and
communicates using the RS-232 serial communications standard~\cite{rs232}. 
The device is controlled 
via a series of commands written to 8 8-bit registers.  These registers control
the desired communications parameters, notably \textit{baud rate, word length, 
parity}, and the number of \textit{stop bits} used when sending or receiving
data.  The device then either listens for incoming data, or the programmer 
writes \textit{data} to the transmit register which is transmitted 
over the serial bus using the configured parameters. Table~\ref{tab:uartouts}
shows the set of possible values for each output state used in our
software model.

\begin{table}
        \centering
        \scriptsize
	\begin{tabular}{ l  l  l }
		\toprule
		Setting & Values & Description \\
		\midrule
		Word Length & 5,6,7,8 bits & Size of the data to send \\
		Baud Rate & Value between 0-115200 & Base wire clock rate \\
		Stop Bits & 1, 1.5, 2 bits & End-of-frame bits \\
		Parity & None,Odd,Even,High,Low & Meaning of parity bits \\
                TX Data & Value 0-255 & Data to transmit \\
		\bottomrule
	\end{tabular}
	\caption{Relevant RS-232 settings with descriptions}
	\label{tab:uartouts}
\end{table}

The 16550 UART peripheral makes a good, complex use case for model 
learning for several reasons:

\begin{itemize}
        \item{\textbf{Relevance}: The 16550 UART is used in many systems today because 
              it is simple enough for even the most basic operating systems to
              control.}
	\item{\textbf{Hidden Registers}: The device has 12 internal registers mapped
              to 8 register addresses. The learner must discover
              how to access all registers before it can change some
              states.}
        \item{\textbf{Output Interdependence}: Certain outputs can only be 
              observed if other outputs are set to specific values. For 
              example, a setting of 1.5 stop bits can only happen if the word
              length is 5.}
	\item{\textbf{Diverse Output Types}: Some observable outputs can take on
              one of a small set of values, but the baud rate is a single value 
              from a set of $2^{16}-1$ possible values, and the data 
              transmitted over the wire is drawn from $2^8$ values.}
	\item{\textbf{Hidden Mathematical Formula}: The baud rate is determined via an
              inherent mathematical formula of 115200 divided by the 
              concatenated value of two other registers.  The learner will have 
              to discover this hidden constant to accurately model the device.}
\end{itemize}

Many devices have traits similar to the above. 
Our simulated 16550 machine simulates register inputs and their
meanings at the bit level.  It simulates data transmission only, read commands
are recognized but return no data, as there is no simulated peer from which 
to receive data.  While limited, this software model is sufficient to 
simulate the interesting complexities of a UART detailed above. If a RNN can 
model a 16550 UART accurately then that is a good indication that it can 
successfully model more complex devices.


\subsection{Generating Observations}

We generated a uniform random set of input sequences for each simulated machine
under test, ran those through each machine and recorded the corresponding 
output set to create a dataset of observations used
to train the RNN.  Random sampling of the input space is not ideal, 
but is sufficient for this experiment as it represents a worst case scenario 
versus a more intelligent sampling scheme. 

\begin{figure}
        \includegraphics[width=\linewidth]{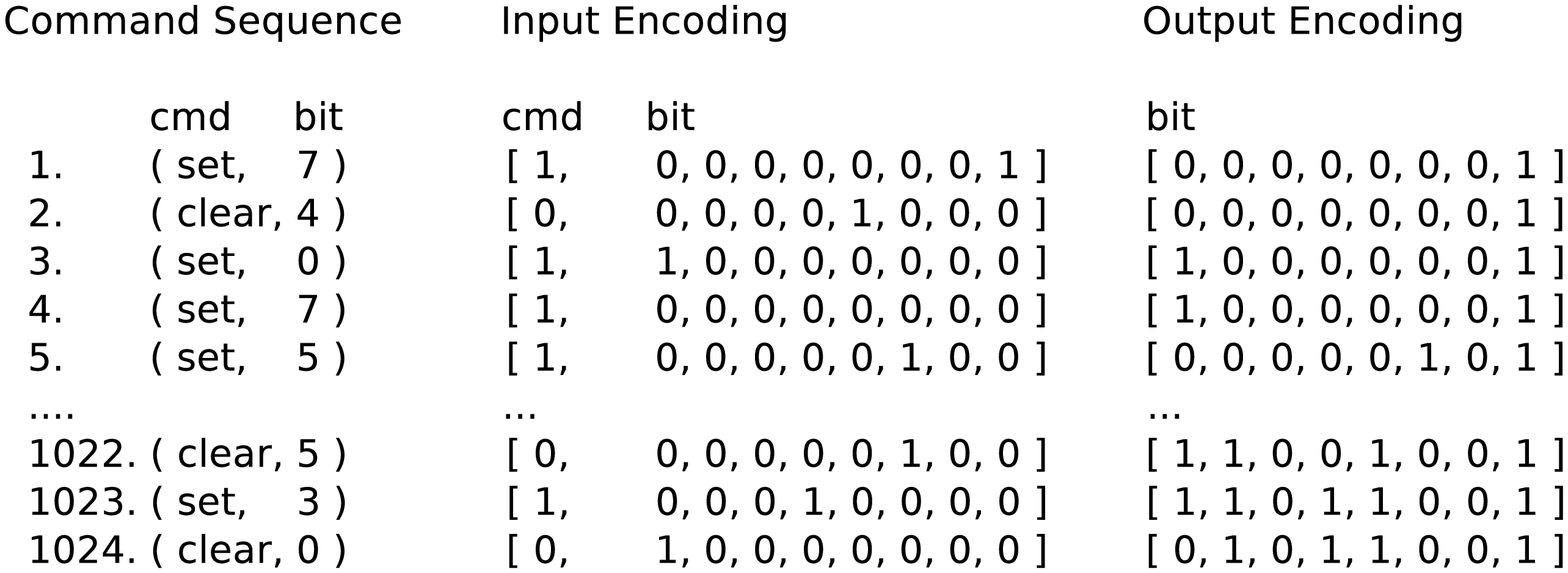}
        \caption{Example command sequence with encoding for EightBitMachine. 
                The input and output for the sequence is encoded as a 
                2-dimensional floating point array.}
        \label{fig:bitmachenc}
\end{figure}

\begin{figure}
        \includegraphics[width=\linewidth]{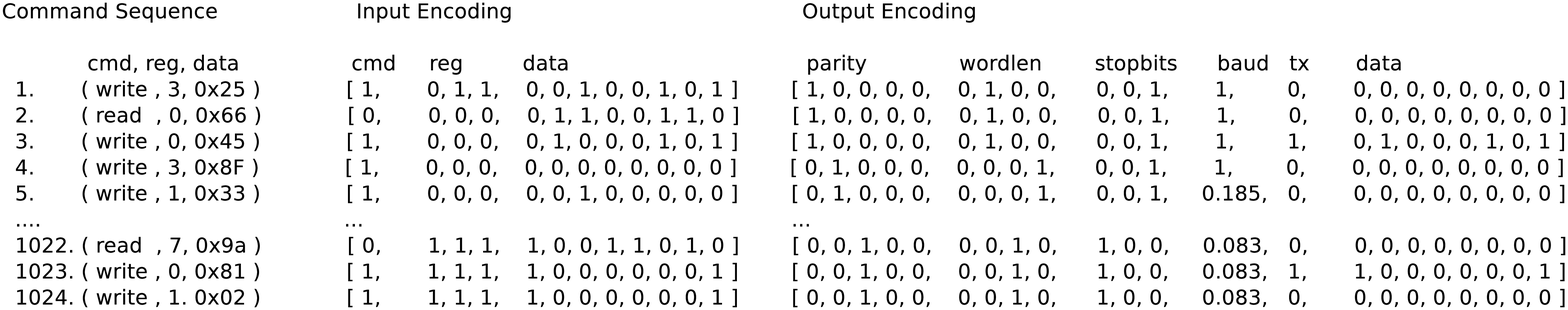}
        \caption{Example command sequence with encoding for SerialPortMachine. 
                Note that parity, word length, and stop bits are one-hot 
                encoded, data is binary encoded, baud rate is linearly scaled
                between 0 and 1, and the transmit flag is True/False.} 
        \label{fig:uartmachenc}
\end{figure}

For the simple machines, the input sequences consisted of a tuple of a 'set' 
or 'clear' command, and the number of the input to modify. The output is the 
vector of results from the machine, with each output bit set to $1.0$ or $0.0$,
representing if the bit was set or cleared. See Figure~\ref{fig:bitmachenc} 
for a detailed example of a command sequence and its input and output encoding.
For the 16550 UART, the input sequences are a tuple of command, register, and 
data; where command is either 'read' or 'write', register is the offset of the 
register to act upon, and data is the 8-bit value to write (it is ignored for 
read commands).  The output of the machine is the state of each output 
specified in Table~\ref{tab:uartouts}, with parity, stop bits, and word length 
encoded as one-hot entries; baud rate encoded as a scaled floating point value 
between $0.0$ and $1.0$; a single output flag that determines whether data was 
transmitted at that time step; and if so, the data transmitted encoded as 8 
binary digits. See Figure~\ref{fig:uartmachenc} for an example encoding of a 
sequence.

Using the above encodings, we generated 4096 training sequences, 1024 
validation sequences used during training, and a separate set of 128 sequences 
used for evaluation after training is complete.  Each training, validation, and
evaluation sequence contains 1024 commands. 

\subsection{Recurrent Neural Network}

\begin{figure}[t]
        \includegraphics[width=\linewidth]{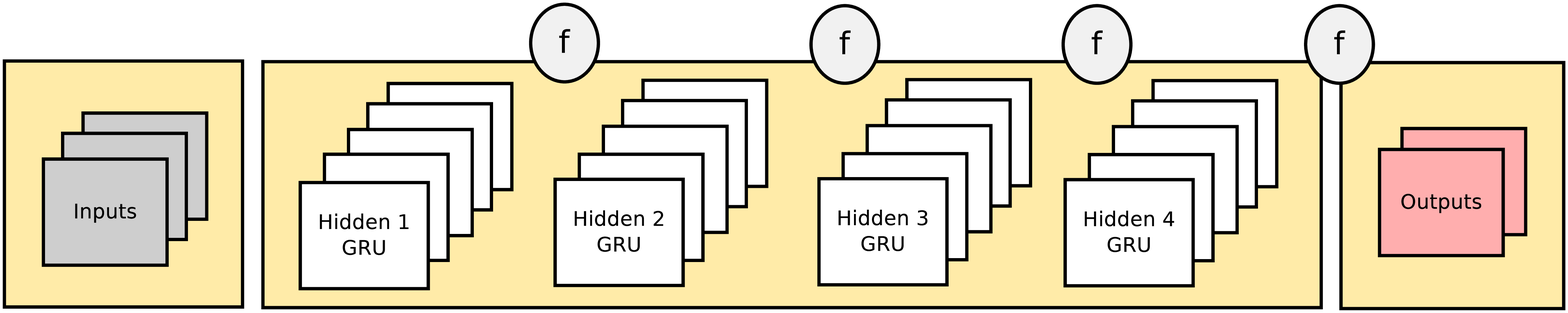}
        \caption{Neural network structure learning device models. The width of 
                each hidden layer is chosen to be larger than the maximum size
                of either the input or output layer.}
        \label{fig:network}
\end{figure}

Our experimental RNNs are multi-layer neural networks consisting of 
GRU~\cite{gru} cells.  Early experiments confirmed that 
non-recurrent networks were unable to learn models of these systems.  Both
GRU and LSTM~\cite{lstm} cells were considered, and while both were able
to learn these models, networks with GRU cells trained faster and were
more stable. The same network structure is used for each test case to make 
sure that the same method will work regardless of the underlying device.  The 
basic structure is shown in Figure~\ref{fig:network}, with four hidden layers 
and an activation function $f$ after each GRU cell. Each layer is fully 
connected to its neighbors. 

Each hidden layer has more cells than the maximum width of either the input
or the output layers. The experiments shown here use the heuristic
$max(I_{width}, O_{width})+1$ to determine the number of recurrent cells in 
each hidden layer. This allows the network to learn the model without
artificial pressure to compress its internal representation.  This heuristic 
does create over-sized networks when there are large disparities between the 
input and output widths which we speculate could lead to overfitting and 
memorization, although that was not observed in these experiments.  

The number of hidden layers was determined experimentally. Early tests showed 
that two hidden layers was sufficient to learn models of all test machines 
except ParityMachine.  Expanding the network to four hidden recurrent layers 
allowed this basic network structure to learn models for all test machines
considered.

\section{Results}
\label{sec:results}

We divide the experimental results into two sets. 
The first set is the  aggregated results of training a large number of networks
for each test machine to verify that learning is taking place. We
compare each network's output to the validation dataset's output, and verify
that the difference between the two, or \textit{validation loss}, is both 
decreasing with training time, and achieves a small value in a reasonable
amount of time.


The second set of results pushes even further and tests whether, with 
continued training, a RNN can exactly model the behavior of a real device
at every time step.  This requires not only achieving a low validation loss,
but also accurate results when presented with previously unknown inputs.  

\subsection{Successful Learning}
\label{subsec:modelres}

We trained 50 recursive networks with the training and validation dataset 
sequences for each machine type using Keras~\cite{keras} with the 
TensorFlow~\cite{tensorflow2015-whitepaper} backend, for a total of 300 
different network instances.  The full set of 
parameters used for training is covered in Appendix~\ref{app:params}. 

Each network was trained until the validation loss for the model dropped 
below $0.1\%$ ($0.001$) for more than 20 consecutive epochs, or a maximum of 
4096 epochs, whichever came first.  When training was complete, each network 
was evaluated further by computing the difference between predicted and
expected output on the previously-unseen evaluation dataset.  The average
evaluation loss among all networks of a particular type is shown in 
Table~\ref{tab:modresults}.  A network is considered ``successfully trained'' 
if the loss on the evaluation dataset is less than $5\%$, a threshold chosen to
show that significant learning had occurred even if the network failed to 
achieve the stopping criteria of $0.1\%$.  Experiments in the
next section will determine if either threshold is sufficient to accurately 
mimic the underlying device's output. By this criteria, the ParityMachine type 
proved hardest to 
learn, with only 78\% of the test networks successfully trained, while all 
other machine types achieved successful learning in $98\%$ of instances or 
higher.  


\begin{table}
        \centering
        \scriptsize
        \begin{tabular}{ l r r r r }
                \toprule
                Machine & \# Params & Epochs & \% Success & Eval Loss \\
                \midrule
                EightBitMachine     & 2,461 & 231  & 98\% & 0.002515 \\
                SingleDirectMachine & 2,461 & 206  & 98\% & 0.003003 \\
                SingleInvertMachine & 2,461 & 37   & 100\% & 0.000018 \\
                SimpleXORMachine    & 2,384 & 155  & 98\% & 0.002466 \\
                ParityMachine       & 2,384 & 1875 & 78\% & 0.023999 \\
                SerialPortMachine   & 10,398 & 773 & 98\% & 0.003326 \\
                \bottomrule
        \end{tabular}
\caption{Results of training 50 networks for each machine}
\label{tab:modresults}
\end{table}

\begin{figure}
        \centering
	\includegraphics[width=0.9\linewidth]{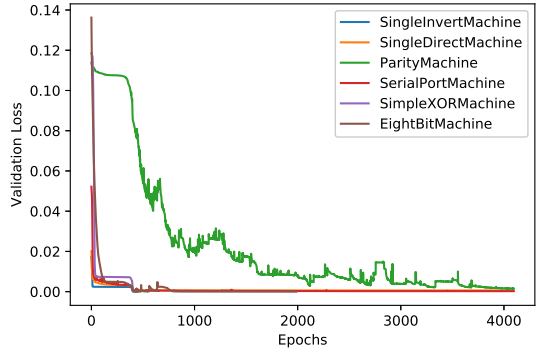}
	\caption{Average validation loss per epoch using only the networks that
                successfully trained for each machine type.}
	\label{fig:valloss}
\end{figure}

Figure~\ref{fig:valloss} shows the average validation loss per epoch 
for each machine type. This shows that most machine types, on average, approach 
our terminal value of $0.1\%$ validation loss within the first 100 
epochs, with the notable exception of ParityMachine, which often takes 
hundreds or thousands of epochs before terminating, if it learns at all. 
The SerialPortMachine type in particular is much more complex than the others
yet still converges quickly, indicating RNNs can model a large 
subset of devices.

\begin{figure}
        \centering
        \includegraphics[width=0.9\linewidth]{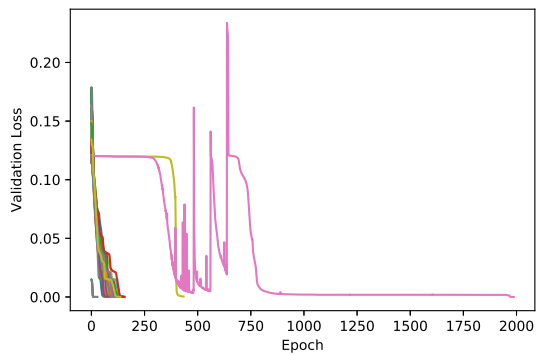}
	\caption{Validation loss for each of 50 trained EightBitMachine networks}
	\label{fig:initexample}
\end{figure}

RNNs are susceptible to unpredictability caused by random weight
initialization.  Despite attempts to control all other variables, we still
observe variability in the success rate and learning performance of each 
individual network. 
For example, Figure~\ref{fig:initexample} shows the 
complete set of successful validation loss performances for each of the 50 
trained networks for the EightBitMachine type.
As shown, the majority of networks started with a loss between $15\%$ and 
$20\%$, and achieved less than $0.1\%$ loss by
epoch 100. There are three outliers; one which starts
off much better and finishes in less than 30 epochs, one that gets stuck around
$12\%$ error until epoch 400, and one that gets stuck at $12\%$ error, and then 
fluctuates wildly until it finally converges near epoch 2000. Similar patterns
re-occur for each machine type.  This indicates it is necessary to train
multiple instances in parallel to guarantee good learning performance.


\subsection{Real-World Effectiveness}
\label{sec:effectiveness}

Having shown that learning is possible for each machine type, 
the next set of results explore how accurately the learned RNN model can mimic 
the behavior of the original device.  This is different than
calculating the global loss over the output sequence because 
the network is returning an encoded floating-point representations of 
predictions on each output using the encoding scheme described in 
Sections~\ref{subsec:simplemachines} and \ref{subsec:16550}.  The predicted
output sequences for each network need to be converted back to their
original output values (via rounding, in most cases) to properly mimic the
output of the original machine.

To evaluate model effectiveness, we choose one of the successful RNNs
from the previous set of results at random, and continue training it until it 
is capable of accurately modelling the evaluation dataset for the underlying 
machine.  This means that every output, at every time step, must be identical 
to that of the original machine.  The results of this experiment are shown in 
Table~\ref{tab:simplemachines}.  The number of outputs 
for each machine indicates the total number of outputs the network must 
correctly predict for 128 evaluation instances, which is output vector width * 
sequence length (1024) * number of instances (128).  The number of extra
epochs of training required to hit the highpoint of accuracy is shown under
``Epochs+'', while ``Epochs'' is the total number of epochs required to 
achieve the result.  Training halted when the network achieved complete 
accuracy on the evaluation dataset for 20 consecutive epochs. 

\begin{table}
        \centering
        \scriptsize
        \begin{tabular}{ l r r r r }
                \toprule
                Machine & \# Outputs & Epochs & Epochs+ & Accuracy\\
                \midrule
                EightBitMachine     & 1048576 & 210 & 57 & 100\% \\
                SingleDirectMachine & 1048576 & 28  & 0 & 100\% \\
                SingleInvertMachine & 1048576 & 69  & 40 & 100\% \\
                ParityMachine       & 131072  & 4026 & 1939 & 99.9992\% \\
                SimpleXORMachine    & 131072  & 83  & 21 & 100\% \\
                SerialPortMachine   & 2883584 & N/A & 33000+ & N/A \\
                \bottomrule
        \end{tabular}
\caption{Maximum evaluation accuracy with output mapping.}
\label{tab:simplemachines}
\end{table}

These results clearly show that a low validation loss in training does not
always translate into perfect real-world performance.  While it is true that
the SingleDirectMachine did not require any more training to achieve perfect
evaluation accuracy, every other machine required more epochs to approach 
that goal.  Two models were unable to achieve 100\% accuracy.  The 
ParityMachine type was able to correctly predict all but one output correctly
with 1939 additional epochs, but was unable to move beyond that value despite
letting the experiment run for several thousand more epochs.  Given the 
observed variability in learning this particular machine, we
expect a 100\% accurate ParityMachine model is possible. 

The SerialPortMachine type was also unable to achieve perfect mimicry by
continuing training on this test, despite having a global error loss 
approaching $2*10^{-5}$.  This model is complex, with 22 values in its output
vector representing 6 different values.  We continued to train the 
SerialPortMachine instance for over 33000 extra instances, but failed to  
achieved perfect accuracy on the evaluation dataset. Table~\ref{tab:realresults}
shows the results of a simple "Hello World!" command sequence at three 
different settings for the network at approximately 33000 epochs. We can
observe that while the Word Length, Parity, and Stop Bits settings are correctly
predicted, the Baud rate and Data outputs are far less accurate. 

\begin{table}
        \centering
        \scriptsize
        \begin{tabular}{ l | r r r r c }
                \toprule
                Target & Baudrate & Wordlen & Parity & Sbits & Output \\
                \midrule
                115200,8n1 & 115285 & 8 & None & 1 & Haxxo Wofxp! \\
                9600,7e1   & 1936  & 7 & Even & 1 & Haxxo Wofxp! \\
                2400,7o2   & 853   & 7 & Odd  & 2 & Haxxo Wofxp! \\
                \bottomrule
        \end{tabular}
\caption{Example "Hello World!" output after 33,000+ epochs, single network}
\label{tab:realresults}
\end{table}

\begin{figure}[t]
        \centering
	\subfloat[16 Epochs, $10\%$ Global Loss\label{fig:uart16}]{
		\includegraphics[width=0.8\linewidth]{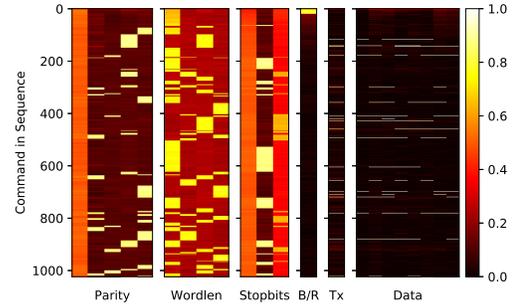}
	}
	\hfill
	\subfloat[176 Epochs, $2.7\%$ Global Loss\label{fig:uard176}]{
		\includegraphics[width=0.8\linewidth]{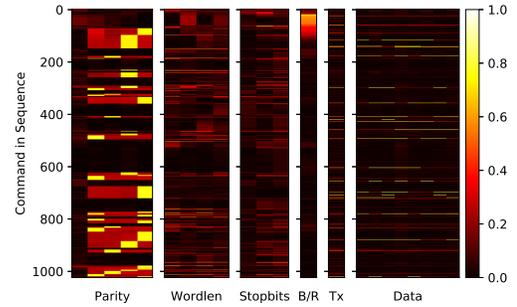}
	}
	\hfill
	\subfloat[2640 Epochs $0.29\%$ Global Loss\label{fig:uart2640}]{
		\includegraphics[width=0.8\linewidth]{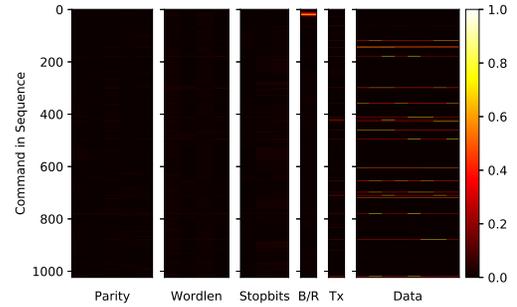}
	}

	\caption{Local error sources contributing to global error for an example
                evaluation sequence of 1024 commands issued to a successful 
                UART model. 
        }
	\label{fig:uartheatmap}
\end{figure}

To determine why, we 
first examine which outputs are contributing the most
to the error. Figure~\ref{fig:uartheatmap} shows validation loss heatmaps of 
each output at different stages of training. Each cell represents the loss of 
a particular output value (x-axis) for each command in 
the 1024-command evaluation sequence (y-axis). Darker colors mean the value at 
that output is closer to the correct value within the sequence.  We show three
maps, one towards the beginning of training, and two more as training 
continues.  We note the evolution of the training over time, and observe the 
overall loss approaching $0.0$. But while the Parity, Word Length, and Stop 
Bits outputs significantly decreased their loss over time, the data 
transmitted and baud rate are the most difficult for the network to 
learn. 

We speculate this is for a few reasons.  First, the baud rate can take
any one of $2^{16}$ possible values, and is encoded as a single 32-bit 
floating point number.  Thus, the network must predict the value to 
within $1.0/2^{16} = 1.5*10^{-5}$ to be correct.  This may be just too 
difficult for the network to do accurately.  Future work will use a 
different encoding for this value. Second, due to randomly generating 
commands for training data, the number of commands in a sequence that 
actually transmit data is small compared to the number of commands overall. 
Thus, we speculate that learning the data outputs is hard due to a 
low proportion of data transmit commands in the command sequence.

\subsection{Decomposed Model}

We speculate a further contributing negative factor to 100\% accuracy with the 
SerialPortMachine is that the value the network is optimizing 
for is the global loss over that instance.  The SerialPortMachine type has 
$22*1024=22528$ outputs per instance, nearly three times as large as any  
other tested machine.  With this large output space, even small amounts of 
noise or error from each cell mask where the network needs 
to minimize the real error.  To test this theory, we created a new model made
up of multiple neural networks, one for each output type: Parity, Word Length,
Stop Bits, Baud Rate, Tx, and Data.  We call this a decomposed model, as it
decomposes the problem into smaller networks. Intuitively, if the network only 
has to optimize for one output type it may be able to learn faster, as it is
optimizing over fewer outputs.

\begin{table}
        \centering
        \scriptsize
        \begin{tabular}{ l l r r r }
                \toprule
                Output & Encoding & Output Size & Epochs & Val. Loss  \\
                \midrule
                Parity & one-hot & 5 & 579 & $5.4*10^{-6}$ \\
                Word Length & one-hot & 4 & 340 & $1.2*10^{-5}$ \\
                Stop Bits & one-hot & 3 & 251 & $2.7*10^{-5}$ \\
                Baud Rate & float & 1 & 1260 & $1.2*10^{-4}$ \\
                Tx & true/false & 1 & 322 & $2.3*10^{-5}$ \\
                Data & binary & 8 & 628 & $1.1*10^{-5}$ \\
                \bottomrule
        \end{tabular}
        \caption{Characteristics of decomposed model networks for UART}
        \label{tab:decomposed}
\end{table}

We are careful to use the same network structure as when 
training with all outputs in one network, and only change the width of the
output layer.   The 16550 UART outputs are not always independent, so the 
network must model all possible states even if it is only predicting one. 
Therefore it is important to keep the same hidden layer widths from the
non-decomposed network to avoid state compression pressure when training 
the decomposed models.  

\begin{table}[t]
        \centering
        \scriptsize
        \begin{tabular}{ l | r r r r c }
                \toprule
                Target & Baudrate & Wordlen & Parity & Sbits & Output \\
                \midrule
                115200,8n1 & 115199 & 8 & None & 1 & H!llo wop,d! \\
                9600,7e1   & 2210  & 7 & Even & 1 & Hello w/rld! \\
                2400,7o2   & 146   & 7 & Odd  & 2 & Hello w/rld! \\
                \bottomrule
        \end{tabular}
        \caption{Example "Hello World!" output, multiple networks}
        \label{tab:realdec}
\end{table}

Metadata about these networks is shown in 
Table~\ref{tab:decomposed}, including how long each took to reach the 
stopping criteria and the validation loss at that time. The result of 
the example ``Hello World'' from before is shown in Table~\ref{tab:realdec}.  
While still not able to achieve 100\% accuracy, the decomposed model is 
significantly better at predicting the final data transmitted, and trains
in much less time than training the entire model at once.

\section{Discussion}
\label{sec:discussion}

These results show that recurrent neural networks can be trained to learn 
information about the inner workings of black box devices.  For the simple 
test machines, the results are very accurate, and while the UART model was 
not able to precisely learn all outputs, it was able to accurately model 
several components of the system and make progress towards the remainder.  
We are confident that perfect accuracy on all models is achievable. 

It is important to note that the structure of the networks used for each 
machine type is identical.  With expert knowledge, one could create network 
structures tailored for a specific target machine, but this generic structure 
shows that modeling can be successful without specific knowledge of the 
underlying machine. Thus, this method is applicable to a wide range of 
use cases.

These results suggest that minor tweaks can be made to the methods presented
here to achieve even more accurate results in the future.  While true, the 
experimental results here are sufficient to support the notion that RNNs are 
useful tools for modeling systems like computer peripherals. 


Furthermore, compared to traditional black box methods like L*, RNNs are able 
to model more complex systems.  State of the art black box systems can only
accurately model devices with a few hundred unique internal states. 
While our simple test machines have up to $2^{8}$ possible internal states, real
devices like the 16550 UART has on the order of $2^{37}$ possible internal
states (See Appendix~\ref{app:statespace}. Such systems are simply too complex 
to learn with traditional black box algorithms. 

The observed difficulty in learning a larger, complex systems like the UART 
suggests that global output loss may not be the optimal parameter to optimise 
for, as the larger the number of outputs and the longer the sequence, the less 
impact each individual output has on that value. Our decomposed model is able 
to overcome this issue for our test case, delivering better results within 
a fraction of the time. 

Finally, we note that the observed validation loss
highlights the difficulty of knowing when to terminate training.  
Some networks had long periods of 
little to no improvement in validation loss, only to suddenly learn the model
hundred or even thousands of epochs later. 

\section{Future Work}

We plan to expand testing to include peripheral 
devices which utilize large memory maps, such as VGA text mode, or DMA and 
interrupts, such as a simple network card.  The UART model is being improved 
to generate and interpret raw RS-232 waveforms to infer settings and data, as 
opposed to the current software model which simply supplies that information 
at each time step.  This leads to interesting learning challenges as the same 
waveform can map to multiple meanings, introducing ambiguity in the training 
data.  This 
will also allow the learned models to interact with physical UART devices.

Further research is needed into output encodings and their impact on learning.
The baud rate, encoded as floating point, appeared to be the hardest for the 
networks to learn, so different encoding techniques will be explored to 
quantify the practical limits on regression accuracy when predicting floating 
point values. 

Future work will also focus on efficient methods to extract the learned 
automata from the neural network.  This will allow automated documentation
and explanation of unknown peripherals to the user. 


\section{Conclusions}

This paper shows empirically that RNNs are capable of modeling even
complex real-world devices accurately using single, generic network 
structure. In addition, we introduced a sample test machine dataset 
useful for evaluating other techniques for modeling peripheral devices.  
With time, we hope the 
technique can be improved and combined with automata extraction to gain
unprecedented insight into the inner workings of unknown peripherals. 

\section*{Acknowledgements}

The author would like to thank Dr. Tim Oates for helping develop 
the ideas in this paper, and Dr. Vincent Weaver for providing computing 
resources for these experiments.

\bibliographystyle{IEEEtran}
\bibliography{IEEEabrv,references.bib}

\appendices
\section{Machine State Spaces}
\label{app:statespace}

Each machine has an input state space, and output state space, and an
internal state space.  The input state space is the number of possible
valid inputs, the output state space is the number of possible valid outputs, 
and the internal state space is the number of possible internal states the
machine can represent at a time.  The magnitude of the state space of each 
machine for a single command is shown in Table~\ref{tab:sspace}.

\begin{table}
        \centering
        \scriptsize
        \begin{tabular}{ l | l l l }
                \toprule
                Machine & Input & Output & Internal \\
                \midrule
                EightBitMachine     & $2^9$    & $2^8$ & $2^8$ \\
                SingleDirectMachine & $2^9$    & $2^1$ & $2^1$ \\
                SingleInvertMachine & $2^9$    & $2^1$ & $2^1$ \\
                SimpleXORMachine    & $2^9$    & $2^1$ & $2^2$ \\
                ParityMachine       & $2^9$    & $2^1$ & $2^8$ \\
                SerialPortMachine   & $2^{12}$ & $2^{37}$ & $2^{37}$ \\
                \bottomrule
        \end{tabular}
\caption{Approximate magnitude of state spaces for each machine.}
\label{tab:sspace}
\end{table}

\section{Keras Parameters}
\label{app:params}

The full list of Keras RNN hyperparameters used for the experiments in this 
paper is shown in Table~\ref{tab:kerasparams}.  A full analysis of how these
parameters were chosen is outside the scope of this paper, but they were 
chosen empirically via sampling the hyperparameter search space. 
Some alternative options are shown in the Options column. 
The hyperparameter that had the most impact on modelling success was the choice 
of activation function.  The experiments in this paper use 
\textit{tanh} as the activation function as it provided the most consistent
and accurate results with a low probability of instability during training.
The \textit{selu} function also performed well. Interestingly, the similar 
\textit{sigmoid} activation function performed poorly in nearly all early 
experiments, as did the \textit{relu} family of functions. Dropout and other 
regularization techniques were not enabled, as over-fitting was not observed 
in these experiments.
%
\begin{table}
        \centering
        \scriptsize
        \begin{tabular}{ l l l }
                \toprule
                Parameter & Value & Options \\
                \midrule
                Layer Type     & GRU & GRU,LSTM,SimpleRNN,Dense \\
                Activation Function & tanh & tanh,linear,(s)elu,sigmoid,relu \\
                Optimizer  & nadam & adam,nadam,rmsprop,sgd \\
                Loss Function & msle & msle,mse,mape \\
                \# Hidden Layers & 4 & 1,2,4 or 8 \\
                
                \bottomrule
        \end{tabular}
\caption{Keras hyper-parameters used for experiments}
\label{tab:kerasparams}
\end{table}

\end{document}